\documentclass[letterpaper]{article}
\usepackage{iccc}

\usepackage{times}
\usepackage{helvet}
\usepackage{courier}
\usepackage{graphicx}
\usepackage{url}

\title{Latent Lab: Large Language Models for Knowledge Exploration}

\newcommand{\cameraready}{1} 

\ifnum\cameraready=1
    \author{Kevin Dunnell\textsuperscript{2}, 
    Trudy Painter\textsuperscript{1}, 
    Andrew Stoddard\textsuperscript{1}, 
    Andy Lippman\textsuperscript{2}\\
    \textsuperscript{1}Department of Electrical Engineering and Computer Science, MIT\\
    \textsuperscript{2}MIT Media Lab\\
    \{tpainter, dunnell, apstodd, lip\}@mit.edu\\
    }
\else
    \author{Anonymous}
\fi
\begin{document} 
\maketitle
\begin{abstract}
\begin{quote}
This paper investigates the potential of AI models, particularly large language models (LLMs), to support knowledge exploration and augment human creativity during ideation. We present ``Latent Lab" an interactive tool for discovering connections among MIT Media Lab research projects, emphasizing ``exploration" over search. The work offers insights into collaborative AI systems by addressing the challenges of organizing, searching, and synthesizing content. In a user study, the tool's success was evaluated based on its ability to introduce users to an unfamiliar knowledge base, ultimately setting the groundwork for the ongoing advancement of human-AI knowledge exploration systems.
\end{quote}
\end{abstract}

\section{Introduction}
The untapped potential of collective knowledge holds significant implications for idea evolution and innovation across various entities \cite{curley2013open}. Despite the digital revolution, information organization remains strikingly similar to traditional methods, limiting exploration across diverse sources and impeding the discovery of interconnected relationships. Current search approaches prioritize quick answers and display results in a list format. This hinders the discovery of interconnected relationships required for meaningful exploration and undermines the context of search terms by prioritizing keywords over semantics.

In contrast, synthesis tools like ChatGPT\footnote{ \url{https://chat.openai.com/}} offer a paradigm shift in user interface design through conversational interaction, though they have drawbacks such as the opaqueness of information sources and limited text-based interaction. This paper outlines the development of Latent Lab\footnote{Try Latent Lab at \url{https://latentlab.ai/}} and evaluates it in the context of the MIT Media Lab data set of 4,000+ research projects. This exploration tool transcends previous search and synthesis tools by incorporating browsing and active visual interaction. Leveraging data manipulation libraries, interactive visuals, and LLMs, Latent Lab overcomes the constraints of keyword-centric search, allowing users to engage in semantically meaningful exploration and synthesis of large data sets. The iterative design process of the tool itself highlights the importance of exploration in the creative process, offering a glimpse into the potential of AI-assisted idea generation.

We make the following contributions to the field of human-AI interactive knowledge exploration systems.
\begin{itemize}
  \item We present the design and implementation of an interactive knowledge visualization tool, including a novel automated technique to label idea clusters using an LLM. 
  \item We report the results from a user evaluation study, demonstrating the utility of a hybrid search/synthesis system to find meaningful insights and connections often overlooked by traditional search and synthesis tools.
\end{itemize}


\begin{figure*}
    \centering
    \includegraphics[width=\textwidth]{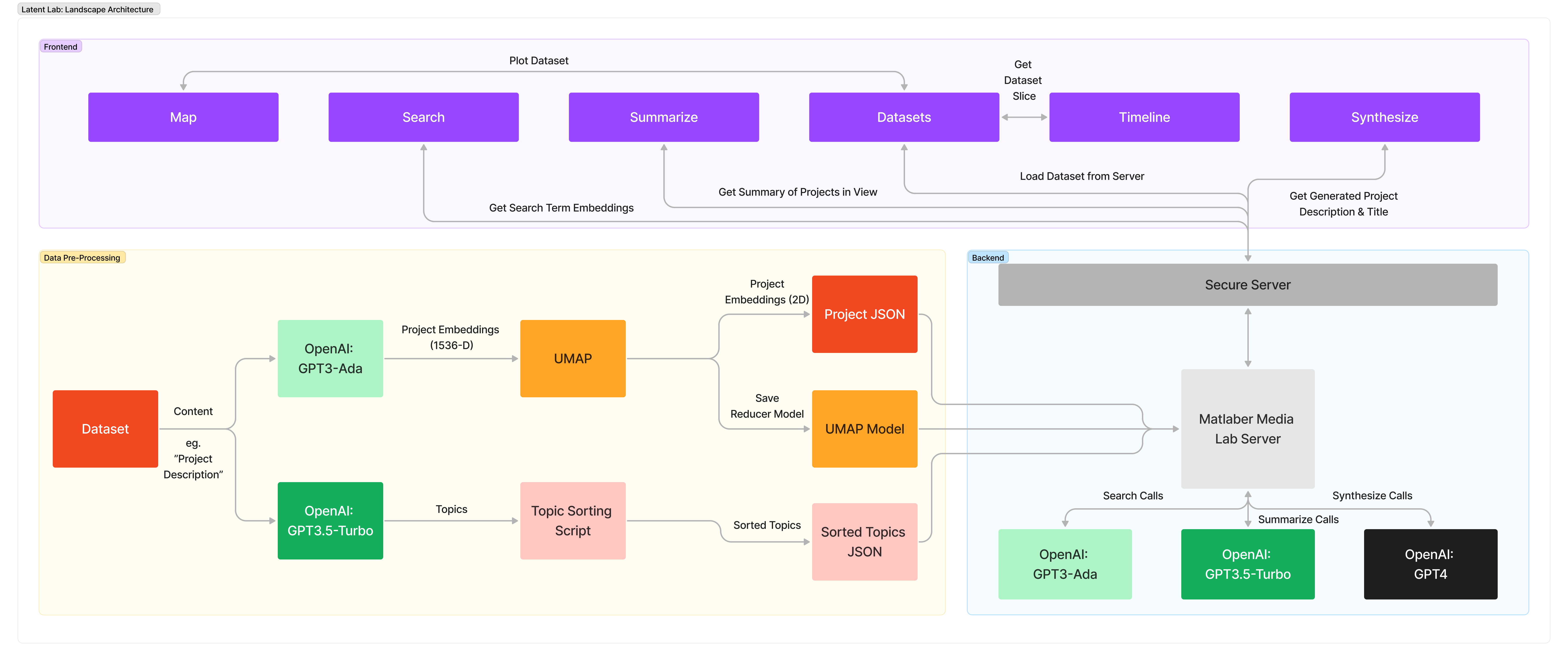}
    \caption{System Architecture of Latent Lab}
\end{figure*}

\section{Related Work}

\subsection{Knowledge Organization}
Vannevar Bush's memex laid the foundation for hypertext and associative indexing \cite{bush1945as}. Richard Feynman's triangulation method emphasized understanding relationships between concepts \cite{feynman_tips_2006}. These ideas influenced the development of Google Knowledge Graph \cite{carr2007freebase}. Our approach to knowledge organization builds on these works to enable fluid exploration of linked information.

\subsection{Information Visualization}
Shneiderman's taxonomy established information visualization principles, with the ``overview, zoom and filter, details-on-demand" mantra guiding the design of visual interfaces for interacting with large data sets. \cite{shneiderman1996taxonomy}. Bostock et al. presented D3.js for interactive visualizations \cite{bostock2011d3}. Heer and Shneiderman highlighted the importance of interaction in visual analysis \cite{heer2012interactive}. Our work integrates these principles to create an informative interface for users.

\subsection{Information Retrieval}
Spärck Jones introduced the term frequency-inverse document frequency (TF-IDF) weighting scheme for keyword-based search \cite{sparck1972statistical}. Mikolov et al. proposed the Word2Vec model for embedding-based search \cite{mikolov2013efficient}. Devlin et al. developed BERT, which further improved semantic search \cite{devlin2018bert}. Latent Lab extends this work, using embedding-based search for relevant results in complex data landscapes

\subsection{Human-AI Collaboration}
Minsky's Society of Mind proposed human intelligence as a result of interacting agents \cite{minsky1988society}. Influential works that consider humans and intelligent systems as interacting agents include TRIZ, Polya's work on invention, and Weis and Jacobson's DELPHI framework \cite{weis2021delphi,polya1945solve,altshuller1999innovation}. Our work further examines human-AI collaboration, aiming to create a system that amplifies human capabilities and positions AI as a ``copilot" rather than an ``autopilot."

\section{Methods}

\subsection{System Overview}
Latent Lab is an AI-powered knowledge exploration system. High-dimensional unstructured data is condensed and visualized in an interactive 2D map. The interface allows users to explore labeled clusters of similar topics, search by semantic context, and synthesize new ideas.

\begin{figure*}[!ht]
    \centering
    \includegraphics[width=0.9\textwidth]{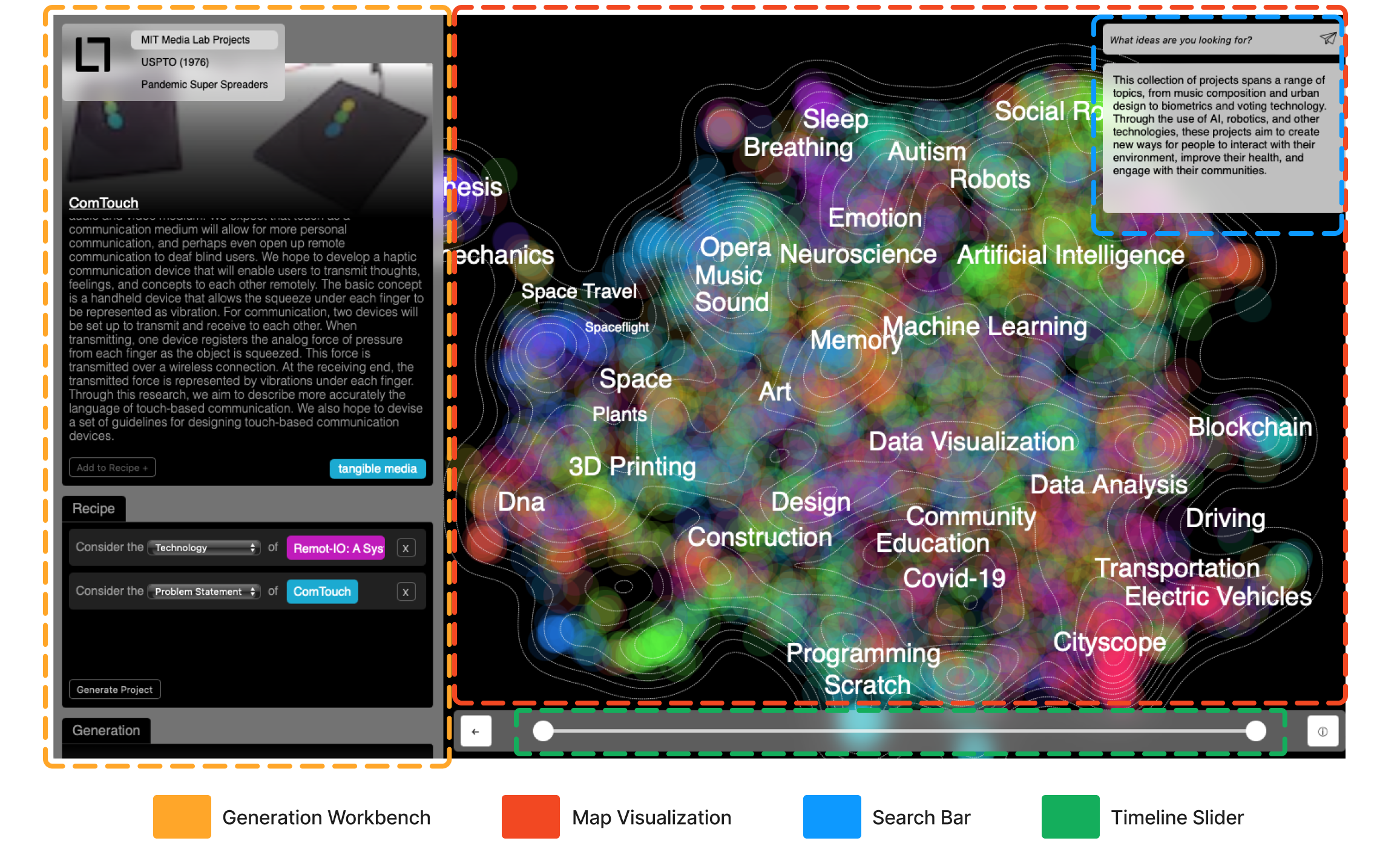}
    \caption{Latent Lab Interface, Annotated to Differentiate Between Components}
\end{figure*}

\subsection{System Architecture}
Latent Lab's system architecture integrates state-of-the-art technologies. The back end is powered by Fast API\footnote{https://fastapi.tiangolo.com/} and Python, while the front is built with Vercel,\footnote{https://vercel.com}, Next.js\footnote{https://nextjs.org}, React\footnote{https://reactjs.org}, and TypeScript. Initially, we aimed to execute all operations on the front end, but the lack of a fully JavaScript-ported version of UMAP \cite{mcinnes2018umap} necessitated the incorporation of a back-end server. This adjustment also enabled server-side rendering, significantly speeding up data loading. The system architecture diagram is presented in Figure 1. 

\subsection{Data Processing}
The data processing pipeline is mostly automated and runs independently of the web app back end for each new data set. It generates three primary artifacts: 

\begin{itemize}
    \item A project JSON containing the unstructured data and embedding data for mapping every project on the front end
    \item A sorted research topics JSON containing all topics produced by the pipeline, ordered by topics with the most associated projects, used for the labels on the front end
    \item A pickled UMAP model to reduce project and topic embeddings to 2 dimensions on the back end
\end{itemize}

\subsection{Topic Extraction}
Latent Lab's automated topic extraction feature sets it apart from other embedding visualization tools, which don't provide insights into cluster meanings. The system uses GPT-3.5-Turbo to distill topics for each project, count occurrences of unique topic labels, and identify related projects. It then calculates label positions using the centroid of the UMAP-reduced coordinates for each associated topic.

\subsection{Components}
The Latent Lab interface has four main components, shown in Figure 2.
It includes a Map Visualization, Generation Workbench, Search Bar, and Timeline Slider.

\subsection{Map Visualization}
The main visualization displays an organized map of project data, with dots representing research projects and clusters indicating semantic similarity. Dot colors correspond to different Media Lab research groups and can be customized to represent other discrete data set attributes.

Contour lines in the map indicate data density within clusters, a concept borrowed from topographic maps where they represent elevation. Paired with the timeline, the changing contour lines reveal the evolution of research concentration. 

Users can pan and zoom, uncovering varying levels of information. High-level labels and contour lines are shown at the highest zoom level, while sub-topic labels and project details appear when zooming in. An occlusion algorithm determines label visibility based on popularity and bounding box overlap.

\subsection{Generation Workbench}
Latent Lab's Generation Workbench allows users to create a ``recipe" for collaboratively synthesizing new research project ideas. Users can choose whole projects or specific aspects, such as community, problem statement, or technology, to include. Once a recipe is prepared, selecting ``generate" submits a preset prompt with selected project elements to GPT-4 via the OpenAI API, producing a synthesized project title and description. Users can view the exact prompt by clicking the ``What was used to generate this?" information button. See Figure 3 for the user flow diagram.

\begin{figure}
    \centering
    \includegraphics[width=\columnwidth]{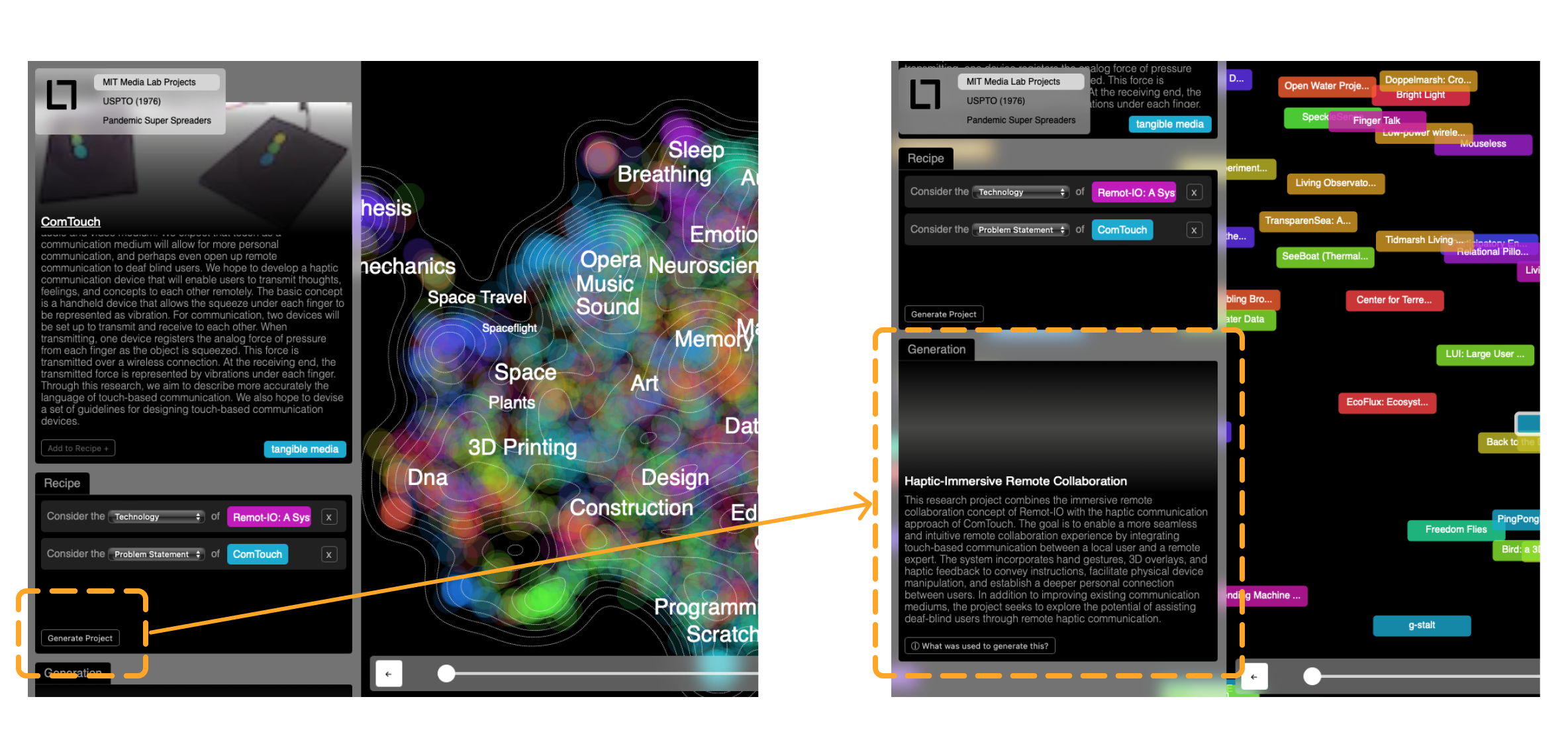}
    \caption{Generating a Research Project Idea}
\end{figure}

\subsection{Search \& Summarization}
Latent Lab employs embedding-based search for semantic meaning instead of simple keyword-matching, enabling more intuitive project exploration through contextual relationships. When a user searches, the query is sent to the back-end server, and the GPT-Ada API returns a 1,536-value embedding. This is passed to the UMAP reducer, yielding x and y coordinates, which are sent to the front end to zoom and highlight the relevant map region dynamically. Figure 4 demonstrates this process using ``quadratic voting" as the search term. Below the search bar, Latent Lab displays summaries that give users a quick overview of projects in the currently viewed map region.

\begin{figure}
    \centering
    \includegraphics[width=\columnwidth]{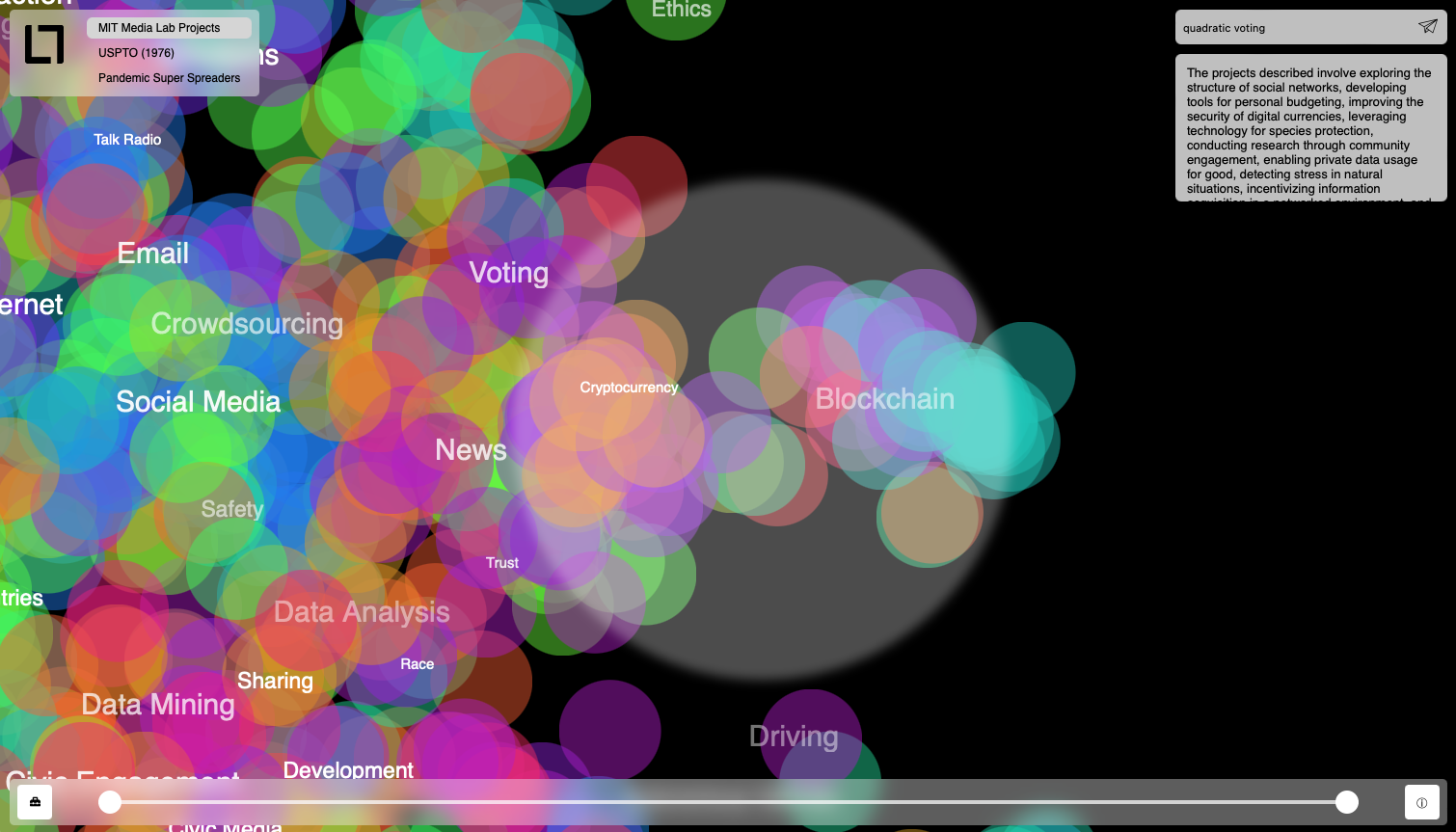}
    \caption{Search Highlighting in Map}
\end{figure}

\subsection{Timeline Slider}
Latent Lab's Timeline Slider enables users to explore data set progression over a selected period using start and end date sliders. This functionality, particularly useful alongside the search bar, allows for efficient examination of current or ongoing projects in specific areas. Figure 5 illustrates timeline filtering for projects since 2018.

\subsection{Study Overview}

We designed a study to evaluate users' experience exploring MIT Media Lab research using Latent Lab. The study compared Latent Lab to the current MIT Media Lab website, which uses traditional keyword-based search. Surveying 94 self-identified researchers via Prolific, participants interacted with both tools in a randomized order. After using each tool, they answered questions assessing clarity, effort \cite{Hart1988DevelopmentON}, engagement, mental support, future use, trust (benevolence, capability, and reliability) \cite{10.2307/258792}, and insight on a 1-5 Likert scale. The study aimed to measure Latent Lab's effectiveness in fostering human-AI collaboration, enhancing user experience, and promoting a deeper understanding of Media Lab projects with AI-powered tools.

\section{Results}

\begin{figure}
    \centering
    \includegraphics[width=\columnwidth]{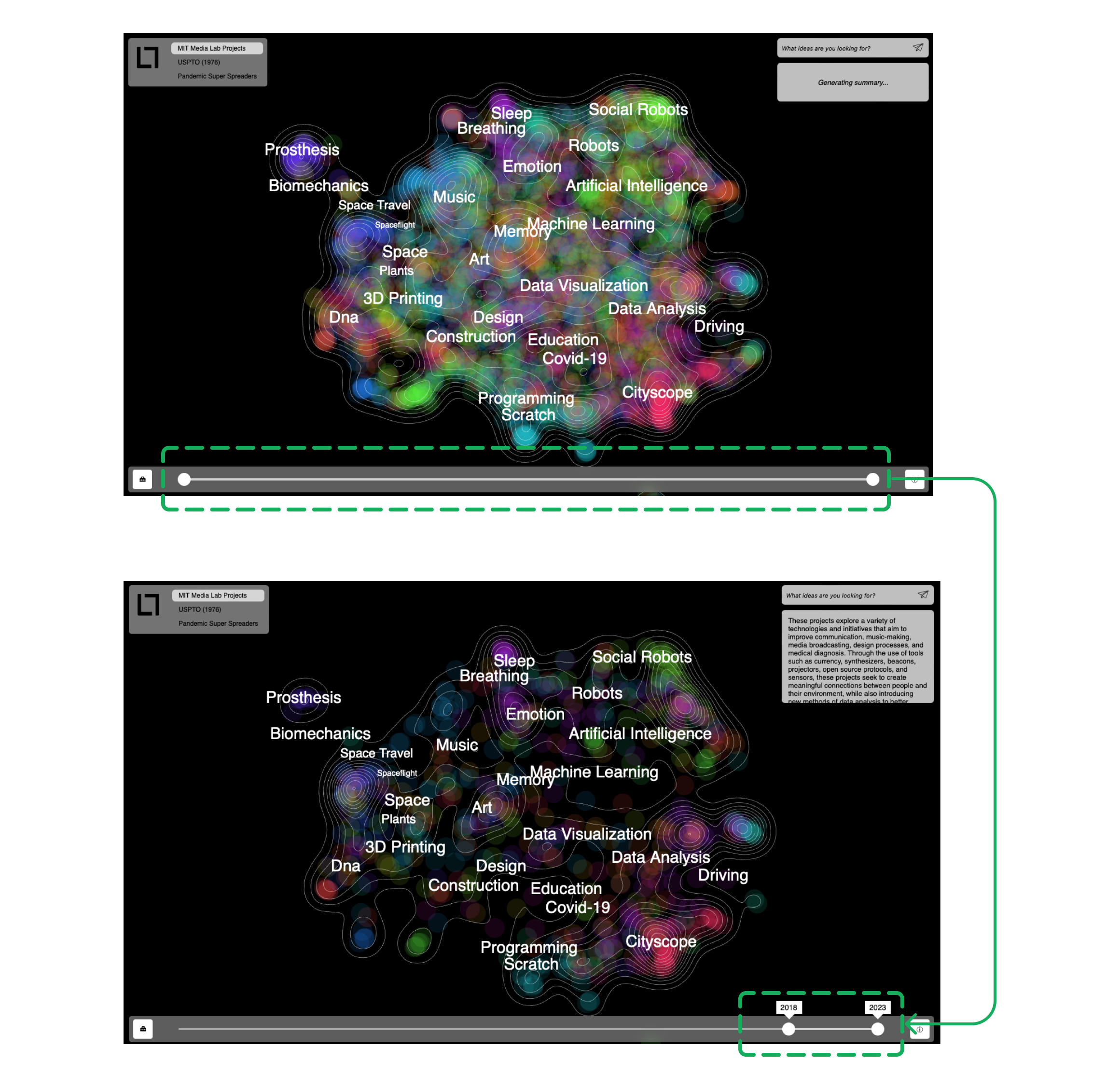}
    \caption{Timeline Evolution}
\end{figure}

\subsection{Analysis}

\begin{figure}
    \centering
    \includegraphics[width=\columnwidth]{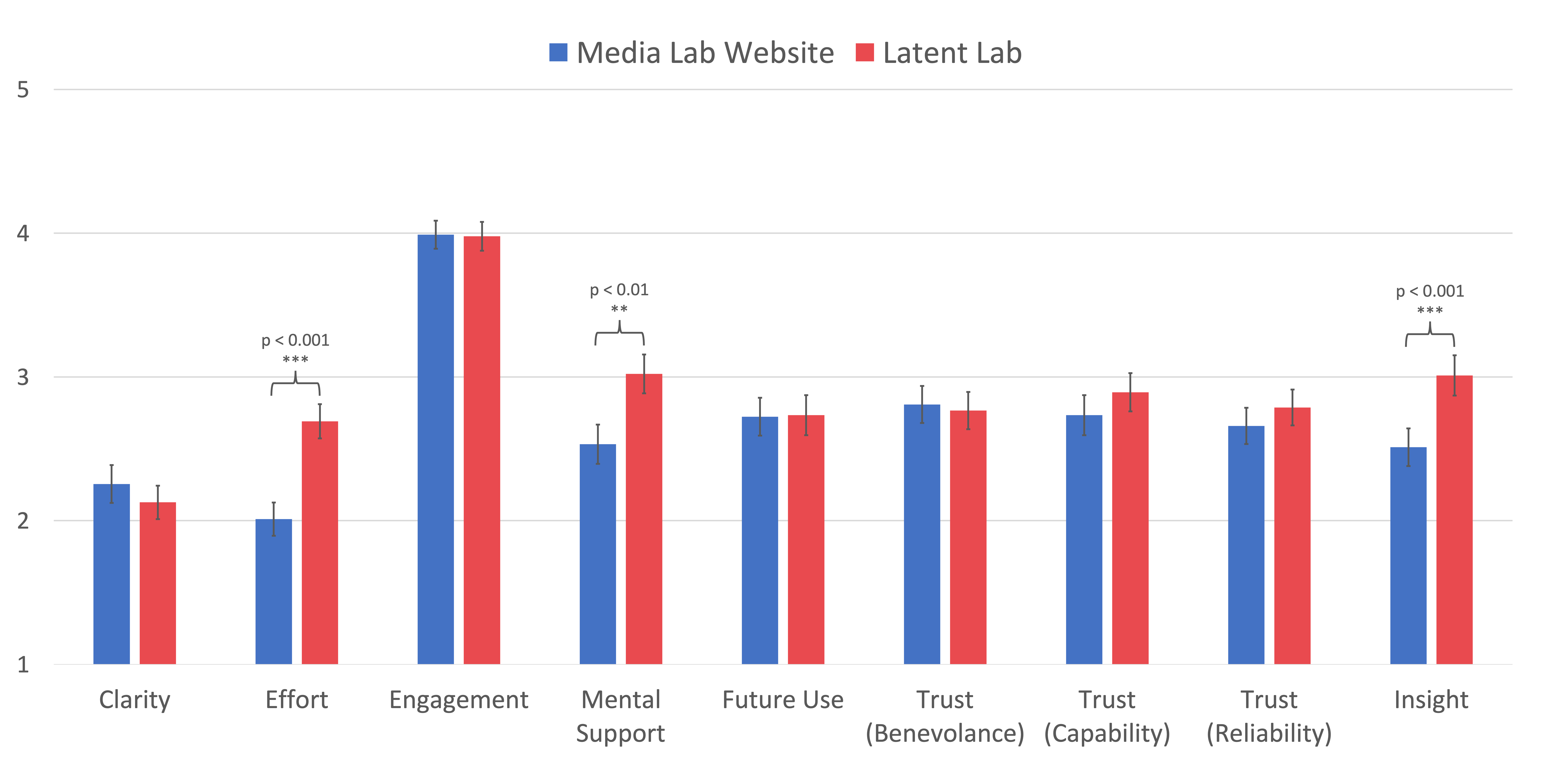}
    \caption{User Evaluation Results}
\end{figure}

The study results in Figure 6 indicate that Latent Lab shows promise as an AI-assisted exploration tool compared to the Media Lab website. Participants were equally engaged, trusted both systems and expressed equal likelihood to use them in the future.

Although Latent Lab required more effort, this is likely due to its novel functionality compared to traditional search interfaces. As users become more familiar with the design, we expect this effort to decrease, facilitating seamless human-AI collaboration. Latent Lab outperformed the Media Lab website in providing higher mental support and insight, suggesting that its semantic map effectively organizes knowledge and offers a deeper understanding of MIT Media Lab research than the current Media Lab website.

Overall, the study highlights Latent Lab's potential and underscores the need for minor improvements to deliver a consistent user experience and enhanced search results.

\subsection{Future Directions}




While our initial inspiration was drawn from our system's ability to generate research project ideas, early user feedback underscored the need to refine Latent Lab's knowledge organization for enhanced exploration, which took precedence over a thorough evaluation of the generated ideas. Looking ahead, our research will adopt a two-fold approach. Firstly, we aim to conduct a comprehensive evaluation of our tool's creativity as an ideation system, benchmarking the utility, novelty, and feasibility of the generated ideas. Secondly, we intend to enhance system performance for handling large user-uploaded datasets and improve data navigation and usability. This will necessitate a focused study on data visualization techniques to optimize Latent Lab's usability, with the ultimate goal of reducing user effort and maximizing the tool's potential for insight extraction.



\section{Conclusion}
Latent Lab serves as an innovative and powerful tool for exploring interconnected relationships within large data sets. By utilizing LLMs and visually engaging interfaces, it transcends conventional search limitations, providing a semantically meaningful and context-aware experience. Emphasizing the value of exploration and iterative design, Latent Lab realizes the long-sought goal of information technology experts for an intuitively accessible wealth of interconnected information. AI-assisted exploration has turned this vision into reality, setting the stage for future human-AI co-invention systems and fostering more intuitive and productive collaborations that are capable of generating novel and impactful creations.

\section{Author Contributions}
KD is lead author, TP is second author and contributed to writing and editing. KD, TP, and AL contributed to the conception of Latent Lab. KD, TP, and AP all contributed to the development of Latent Lab.

\section{Acknowledgments}
We thank Joshua Bello, Deborah Jang, Michael Peng, Jinhee Won, and Anushka Nair for their help with Latent Lab development. We also thank Ziv Epstein and Hope Schroeder for their guidance with user testing. Finally, we thank all survey participants for their time.

\bibliographystyle{iccc}
\bibliography{iccc}

\end{document}